\documentclass[a4paper]{article}

\usepackage{graphicx}
\usepackage{amssymb,amsmath}

\begin{document}

\title{Spiking Networks for Improved Cognitive Abilities of Edge Computing Devices}
%\subtitle{Concept Paper}

\author{Anton Akusok$^1$, Kaj-Mikael Bj\"ork$^2$, Leonardo Espinosa Leal$^3$,\\
Yoan Miche$^4$, Renjie Hu$^5$ and Amaury Lendasse$^6$}
\date{%
    $^1$Arcada University of Applied Sciences,\\ Jan-Magnus Janssonin aukio 1, Helsinki, Finland\\
    $^2$Hanken School of Economics, Arkadiankatu 22, Helsinki, Finland\\
    $^3$Risklab at Arcada UAS,\\ Jan-Magnus Janssonin aukio 1, Helsinki, Finland\\
    $^4$Nokia Bell Labs, Karaportti 3, Espoo, Finland\\
    $^5$The University of Iowa, 3131 Seamans Center for the Engineering Arts and Sciences, Iowa City, IA, USA\\
    $^6$University of Houston, 4730 Calhoun Road, Houston, TX, USA
}

\maketitle

\begin{abstract}
    This concept paper highlights a recently opened opportunity for large scale analytical algorithms to be trained directly on edge devices. Such approach is a response to the arising need of processing data generated by natural person (a human being), also known as personal data.
    Spiking Neural networks are the core method behind it: suitable for a low latency energy-constrained hardware, enabling local training or re-training, while not taking advantage of scalability available in the Cloud.
\end{abstract}

\section{Introduction}

A sudden realization came to our minds while preparing this white paper -- mobile phones are the first type of devices that received dedicated math accelerators at a pervasive scale. Such things never got wide adoption before: Intel 8087 co-processor\cite{palmer1980intel}, Intel Xeon Phi\cite{chrysos2014intel,jeffers2013intel} or Google TPU (Tensor Processing Unit)\cite{jouppi2017datacenter} stayed niche devices that few people use and even fewer develop for. But since the last two years, major mobile phone companies include dedicated co-processors\cite{mit2018ondevice} necessary for computational photography enhancement or facial recognition, that are suitable for general machine learning.

Currently the dominant analytical approach stores data and runs computations in the Cloud\cite{qiu2018towards}. However Cloud based methods poorly fit to a range of important practical applications including augmented reality, real-time data analysis, real-time user interaction, or processing sensitive data that incur high risks for a company if leaked, stolen or intercepted in transfer. The price of deployed analytical methods is increased by the need to have a permanently working internet connection for users, and cloud hardware rent for service providers.

\subsection{Mobile-first Machine Learning}

The difference in the operating systems running on mobile and desktop devices is smaller than many people think. It is limited to the interface, that interacts with window-based applications using mouse and keyboard on desktops, and with full-screen applications using touch gestures on mobile devices. The operating system kernel, storage and graphics are almost identical to the desktop analogues.

However, there is a factor that amplifies the small differences of mobile devices to the extent of making them useless in machine learning: the software. Researchers use programming languages (Python, R) created for systems with global folder access and command-line installers. These are not readily available on mobile devices that sandbox all applications for security reasons\cite{li2013mobile}. Even if they manage to run on mobile devices, their speed and efficiency are abysmal because they can only access a low-power CPU that needs to emulate advanced x86 architecture commands\cite{hillar2011intel}.

The missing software problem is recognized by the mobile device vendors, who addressed it by providing optimized libraries or even whole new programming languages\footnote{https://swift.org/} that enable writing powerful software for the mobile platform. The authors worked with GPU-accelerated mobile BLAS/LAPACK library\cite{akusok2018high} and found it to be as convenient as high-level CUDA primitives while performing on par at the level of a 45W Intel i7 laptop processor.

\subsection{Spiking Neural Networks as the "mobile brains"}

A good supervised learning method is needed for the mobile-first machine learning, and it must be able to train directly on device. Apple ecosystem provides accelerated libraries for all kinds of linear systems (dense and sparse), and for inference in arbitrary neural networks.

Deep Learning and traditional neural networks can be rejected for their slow and energy-inefficient training. Support Vector Machines and Nearest Neighbour method also slow at large scale. Decision Trees and Random Forest are missing the corresponding libraries, and linear models are limited in their learning ability. Native support for the concept of \emph{time} is another requirement for a human-friendly method operated on mobile devices by actual people, and it is not available in any of the mentioned methods. A recurrent deep neural network with the training speed of a linear model would fit perfectly for mobile devices.

Such method actually exist, in form of spiking neural network\cite{maass1997networks} implemented in reservoir computing framework\cite{schrauwen2007overview}, also known as Liquid State Machine\cite{maass2002real,LSM}. Spiking neurons have an internal state value, that produces a binary output ("spike") and resets to zero upon reaching a threshold. That internal state provides networks of spiking neurons with the concept of time, and binary output spikes are propagated energy efficiently at the lowest possible precision in edge devices. Due to internal states spiking networks can be run only one step at a time, limiting speedup from batch processing on large parallel devices (server-grade GPU accelerators) but fitting well to the low latency (due to local processing) and shared-memory GPUs of edge devices. 

\begin{figure}
    \centering
    \includegraphics[width=0.8\textwidth]{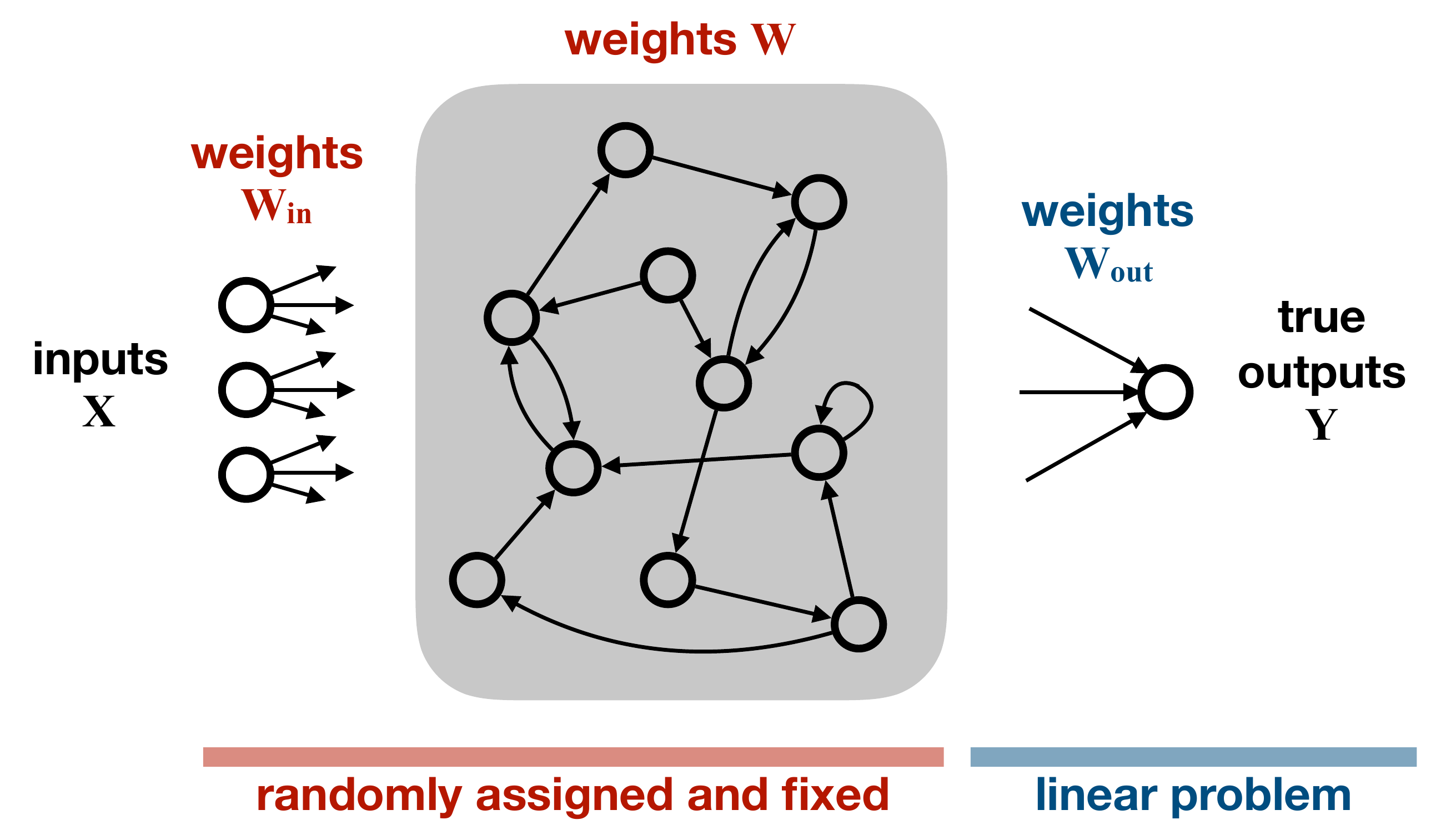}
    \caption{Liquid State Machine: a spiking neural network running in reservoir computing framework. Reservoir part (\emph{gray}) requires only inference; the training occurs in the linear readout layer (\emph{blue}).}
    \label{fig:lsm}
\end{figure}

Spiking neurons have no effective training methods as they are not differentiable; but gradient-based training is a poor choice for edge devices due to high computational power and energy demands. Liquid State Machines (LSM) offer a framework for spiking networks that avoids training the spiking neuron reservoir (a sparsely connected pool of neurons), see Figure~\ref{fig:lsm}. A network is initiated by randomly generating and fixing sparse input weights $\mathbf{W}_\text{in}$, sparse binary connection weights $\mathbf{W}$, and spiking neuron parameters. Inference step consists of propagating spikes inside the reservoir by multiplying neuron outputs with $\mathbf{W}$, compute spiking neurons inputs by adding input vector $X$ multiplied by $\mathbf{W}_\text{in}$, then updating and recording states of spiking neurons. The most computationally demanding part of sparse matrix-vector multiplication with $\mathbf{W}$ is done in lowest precision as the values are binary. The network is trained by learning weights of a linear output layer $\mathbf{W}_\text{out}$ between the recorded spiking neuron states and the true outputs $Y$. Network can be re-trained any time with new true outputs and the recorded states, without the need to re-run inference as all inference parameters are fixed.

\section{Conclusions: Native machine learning at the edge}

The development tools and hardware accelerators for mobile machine learning are already available, but researchers are generally unaware of the computational power in these devices, or are skeptical about feasibility of typical machine learning methods running at edge devices. 

This paper explores the motivation behind training on edge devices, and highlights a suitable method based on spiking neural networks. It fits to the hardware specifics of such devices, less affected by their drawbacks, and makes large-scale machine learning with model training or re-training feasible directly on the edge.

\bibliographystyle{plain}
\bibliography{references}

\end{document}